\pdfoutput=1
\documentclass[10pt,twocolumn,letterpaper]{article}
\usepackage{cvpr}
\usepackage{times}
\usepackage{epsfig}
\usepackage{graphicx}
\usepackage{amsmath}
\usepackage{amssymb}
\usepackage{soul}
\usepackage{bbm}
\usepackage{subcaption}
\usepackage{authblk}
\usepackage{multicol}
\usepackage{float}

\usepackage{mathtools}
\usepackage{wrapfig,lipsum,booktabs}
\usepackage{enumitem}

\newcommand{\ra}[1]{\renewcommand{\arraystretch}{#1}}
\newcommand\blfootnote[1]{%
  \begingroup
  \renewcommand\thefootnote{}\footnote{#1}%
  \addtocounter{footnote}{-1}%
  \endgroup
}

\renewcommand{\thefootnote}{\fnsymbol{footnote}}

\usepackage[pagebackref=true,breaklinks=true,letterpaper=true,colorlinks,bookmarks=false]{hyperref}

\cvprfinalcopy 

\begin{document}

\title{InstantBooth: Personalized Text-to-Image Generation without Test-Time Finetuning}

\author{Jing Shi$^{*}$\quad Wei Xiong$^{*}$\quad Zhe Lin\quad Hyun Joon Jung\quad\\
Adobe Inc. \\
{\tt\small \{jingshi,wxiong,zlin,hjung\}@adobe.com}
}

\twocolumn[{%
\renewcommand\twocolumn[1][]{#1}%
\maketitle
\begin{center}
    \centering
    \captionsetup{type=figure}
    \vspace{-0.2in}
    \centerline{\includegraphics[width=\textwidth]{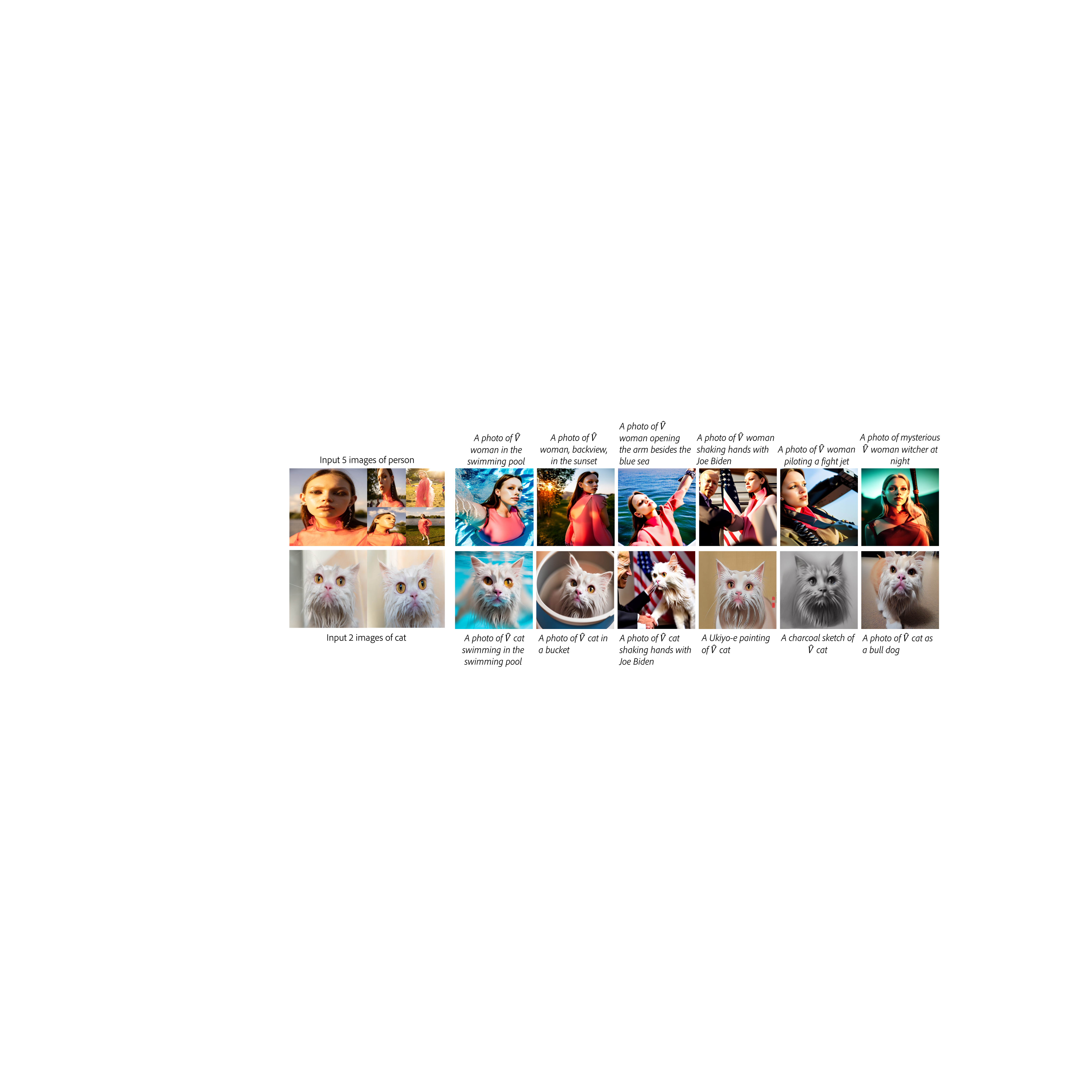}}
	\caption{Personalized images generated by our model given different prompts. On both the ``person'' and ``cat'' categories, our model can generate text-aligned, identity-preserved and high-fidelity image variations of the input concept. $\hat{V}$ is an identifier representing the input subject. \textit{Our model can instantly generate personalized images \textbf{with only a single forward pass}.}}
	\label{fig:teasing}
\end{center}%
}]

\blfootnote{* Equal Contribution}

\begin{abstract}
Recent advances in personalized image generation allow a pre-trained text-to-image model to learn a new concept from a set of images. However, existing personalization approaches usually require heavy test-time finetuning for each concept, which is time-consuming and difficult to scale. We propose InstantBooth, a novel approach built upon pre-trained text-to-image models that enables instant text-guided image personalization without any test-time finetuning. We achieve this with several major components. First, we learn the general concept of the input images by converting them to a textual token with a learnable image encoder. Second, to keep the fine details of the identity, we learn rich visual feature representation by introducing a few adapter layers to the pre-trained model. We train our components only on text-image pairs without using paired images of the same concept. Compared to test-time finetuning-based methods like DreamBooth and Textual-Inversion, our model can generate competitive results on unseen concepts concerning language-image alignment, image fidelity, and identity preservation while being 100 times faster. Project Page: \url{https://jshi31.github.io/InstantBooth/}
\end{abstract}

\section{Introduction}
The goal of personalized text-to-image generation~\cite{ruiz2022dreambooth,gal2022image,kumari2022multi,gal2023designing} is to learn a concept from a set of images, then generate new scenes or styles of the concept from input prompts. As shown in Fig. \ref{fig:teasing}, given a set of images of the same person, personalized image synthesis aims to generate new images of the person with different poses, backgrounds, object locations, dressing, lighting, styles while keeping the same identity. 

Such a problem can be seen as an instance-conditioned generation task with identity preservation and language control. Existing approaches for this task~\cite{ruiz2022dreambooth,brooks2022instructpix2pix,zhang2023adding} are usually built upon pre-trained text-to-image models~\cite{rombach2022high,saharia2022photorealistic} to fully utilize the image generation capacity of big models. These approaches can be categorized into two primary directions. 

The first direction is to invert the input images to the textual space so that the pre-trained models have a deep understanding of the concept. For example, DreamBooth~\cite{ruiz2022dreambooth} learns a  unique identifier from the specific subject by finetuning the whole diffusion model. Textual-Inversion~\cite{gal2022image} inverts the input images to a unique textual embedding and learns the embedding-image mapping during finetuning. During inference, the finetuned model can generate identity-preserved scene variations of the input concept using prompts containing the learned textual identifier. However, the pre-trained model needs to be finetuned for many steps to learn each new concept. The learned model weights need to be stored per concept. It is both time and storage-consuming, and thus the scalability of these approaches is greatly limited.  

The second direction is to learn an image-to-image mapping with text guidance directly. Training such image-conditioned models usually requires large-scale paired training data, i.e., image pairs of the same subject but with different poses or other attribute variations. Otherwise, the language understanding and generalization ability can be forgotten from the pre-trained model if we finetune the model with  limited data. However, such large-scale training pairs are very difficult to obtain. Although there are few explorations on extending pre-trained text-to-image models to take image conditions, their limitations cannot be ignored when applied to personalized image generation. For example, InstructPi2Pix~\cite{brooks2022instructpix2pix} achieves text-guided image-to-image translation by learning all the parameters of a pre-trained text-to-image model with paired images (before and after modification) generated using Prompt-to-Prompt~\cite{hertz2022prompt}. However, it can primarily perform pixel-to-pixel mapping while failing to generate objects with large pose variations or change the location of objects. ControlNet~\cite{zhang2023adding} extends the pre-trained model to take multi-modal data as inputs, but it does not show evidence of identity preservation ability when conditioned on images.

This paper aims at \textit{addressing these challenges by lifting test-time finetuning for personalized image generation}. We achieve this goal with several novel components. Instead of inverting the input images to a token with inefficient online optimization, we propose to learn the general concept of the input images by learning an image encoder and map them to a compact textual embedding. Such an ``offline'' training-based design makes our model generalizable to unseen concepts. However, merely using the compact embedding to represent the concept leads to a lack of fine-grained identity details in the generated images. Inspired by recent works for language and vision model pre-training~\cite{alayrac2022flamingo,li2023gligen}, we introduce a few trainable adapter layers 
 to extract rich identity information from the input images and inject them into the fixed backbone of the pre-trained model. By using the adapter, our model successfully preserves the identity details of the input concept while keeping the generation ability and language controllability of the pre-trained model.

Our new components are trained on only text-image pairs without using any paired images of the same concept. We observe that with such training strategy, our model can generalize well to new concepts. i.e., given images of a new concept, our model can generate objects of that concept with large pose and location variations, satisfactory identity preservation and language-image alignment, as demonstrated in Fig. \ref{fig:teasing}. 

In summary, our contributions are three-fold: 
\begin{itemize}[noitemsep,nolistsep]
    \item This paper is among the first few works that achieve personalized text-to-image generation without test-time finetuning.
    \item We design a new architecture to convert the input images to a textual embedding so that the model is generalizable to unseen concepts. We further inject rich visual feature representation into the pretrained text-to-image generation model to preserve the input identity without losing language controllability.
    \item We demonstrate that our method can achieve comparable results as test-time finetuning-based methods like Dreambooth while being x100 faster.
\end{itemize}

\section{Related Work}
\noindent\textbf{Text-to-Image Synthesis.}
Generative Adversarial Networks~\cite{reed2016generative,xu2018attngan,abdal2022clip2stylegan,gal2022stylegan} are usually used for text-to-image generation in early years. However, they are limited to generating well-aligned and highly-structured objects, such as face and church. More following work are proposed using transformers~\cite{crowson2022vqgan,ding2022cogview2} or diffusion models~\cite{ho2020denoising,avrahami2022blended,bar2022text2live,hertz2022prompt,kim2022diffusionclip,liu2023more,nichol2021glide} to generate high-fidelity images from text. 

Recently, big generative models have emerged to achieve state-of-the-art performance on generating unaligned complex objects. Among them, DALL-E~\cite{ramesh2021zero} employs auto-regressive models to firstly demonstrate that high-quality images can be synthesized using large-scale training. Parti~\cite{yu2022scaling}, a more advanced auto-regressive model, shows that better visual quality and text-image alignment can be achieved by scaling up the model and data. DALL-E2~\cite{ramesh2022hierarchical}, Imagen~\cite{saharia2022photorealistic}, and Stable Diffusion~\cite{rombach2022high}, trained on large-scale text-image pairs, demonstrate the power of diffusion models on generating structures and semantics of complex scenes. eDiff-I~\cite{balaji2022ediffi} improves the text-image alignment of diffusion-based models by combining different pre-trained language models. Muse~\cite{chang2023muse}, a text-to-image Transformer model, proves that beyond auto-regressive and diffusion models, masked generative modeling can also achieve state-of-the-art performance providing enough model capacity and large-scale data. Our model is build upon pre-trained text-to-image diffusion models to fully utilize their generation ability for personalized image synthesis. 

\noindent\textbf{Multimodal-Conditional Image Synthesis.}
Modern visual synthesis systems are usually powered by large text-to-image foundation models. Beyond text guidance, these models can also take additional inputs from other modalities, such as image, depth, layout, and so on. For example, Stable Diffusion V2~\cite{rombach2022high} can either take a depth image as condition to generate images or take an image and a mask for image inpainting. eDiff-I~\cite{balaji2022ediffi} takes images as additional style guidance for stylized image synthesis.

More recently, approaches built upon pre-trained text-to-image models have been proposed for more controllable multi-modal image generation.  ControlNet~\cite{zhang2023adding} controls pre-trained diffusion models to take more input conditions by locking the original parameters and making a trainable copy to the newly added layers. GLIGEN~\cite{li2023gligen} extends pre-trained models to take layouts, images or other conditions by injecting new layers to the original model.
 The goal of our model is similar in that our model aims to extend and control a pre-trained model to take images of a concept as an additional condition for personalized image generation, while fully utilizing the generation ability of the original model.

\noindent\textbf{Personalized Image Synthesis.}
Given one or more images of a concept as inputs, personalized image generation aims at generating image variations of the given concept or identity. 
Early studies in this area usually adopt GAN-based models to generate style variations of the same aligned face or bedroom~\cite{richardson2021encoding}. Recently, several approaches based on pretrained text-to-image models have been proposed~\cite{gal2022image,ruiz2022dreambooth,kumari2022multi}. 
These models take a set of images of a concept, and generate variations of the concept with different text prompts containing this concept.
DreamBooth~\cite{ruiz2022dreambooth} and Textual Inversion~\cite{gal2022image} generate objects with large pose and location variations of the given concept. 
\cite{voynov2023p+} improves Textual Inversion by designing a richer inversion space.
DreamArtist~\cite{dong2022dreamartist} reduces the number of input images to single.
However, for each new concept, they need to finetune the pre-trained model for many steps, which is very time-consuming and thus their application scenario is limited. 
To speed up, CustomDiffusion~\cite{kumari2022multi} and SVDiff~\cite{han2023svdiff} reduce the amount of fientuned parameters  and \cite{gal2023designing} learns an encoder as a good parameter initialization for finetuning. 

Nevertheless, these methods still rely on test-time finetuning.
In contrast, once trained, our approach does not need any test-time finetuning for each new concept and can instantly generate identity-preserved results, which is a significant improvement on efficiency.

We also recognize several concurrent works \cite{wei2023elite,ma2023unified,chen2023subject}. 
To allow the pretrained text-to-image generator to adapt visual input, ELITE~\cite{wei2023elite} finetunes the pretrained parameters in the attention layers and UMM-Diffusion~\cite{ma2023unified} only learns a visual mapping layer but freeze the weight of the pretrained generator. SuTI~\cite{chen2023subject} achieves personalized image generation without test-time finetuning by learning from massive amount of paired images generated by subject-driven expert models. Different from these methods, our method adopts the adapter structure to tightly incorporate the visual signal to the generator structure while freezing the pretrained parameters. Moreover, we do not use paired images to train our model. 

\begin{figure*}[t]
    \centering
    \includegraphics[width=0.95\textwidth]{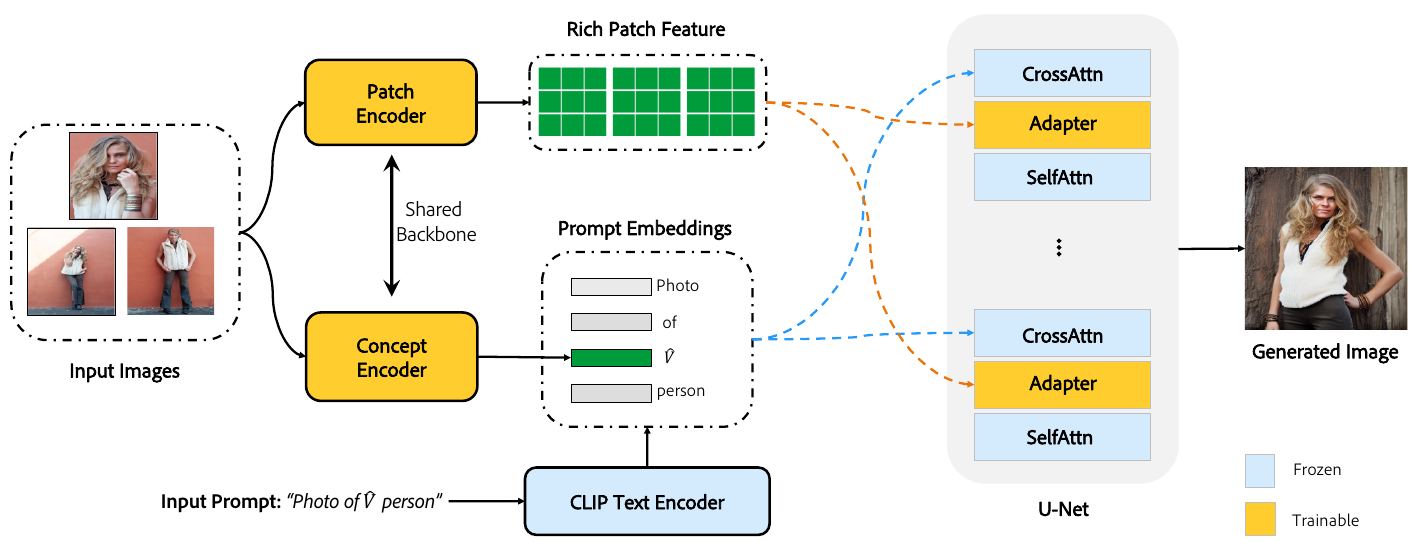}
    \caption{An overview of our approach. We first inject a unique identifier $\hat{V}$ to  the original input prompt to obtain ``Photo of $\hat{V}$ person'', where $\hat{V}$ represents the input concept. Then we use the concept image encoder to convert the input images to a compact textual embedding and use a frozen Text encoder to map the other words to form the final prompt embeddings. We extract rich patch feature tokens from the input images with a patch encoder and then inject them to the adapter layers of the U-Net for better identity preservation.  The U-Net of the pre-trained diffusion model takes the prompt embeddings and the rich visual feature as conditions to generate new images of the input concept. During training, only the image encoders and the adapter layers are trainable, the other parts are frozen. The model is optimized with only the denoising loss of the diffusion model. (We omit the object masks of the input images for simplicity.)}

    \label{fig:framework}
\end{figure*}

\section{Method}
\subsection{Overall Framework}
Given a few images of a concept, the goal is to generate new high-quality images of this concept from text description $p$. The generated image variations should preserve the identity of the input concept. As DreamBooth~\cite{ruiz2022dreambooth} summarized, the variations include changing the concept's location, property or style, modifying the subject's pose, structure, expression or material, \etc.

The overall framework of our model is shown in Fig. \ref{fig:framework}. Our model is built upon a pre-trained text-to-image model. We first inject a unique identifier $\hat{V}$ to the input prompt to represent the object concept, then use a learnable image encoder to map the input images to a concept textual embedding. The pre-trained diffusion model takes the concept embedding along with the embedding of the original prompts to generate new images of the input concept. To enhance the identity of the generated images, we introduce adapter layers to the pre-trained model to take rich patch features extracted from the input images for better identity preservation. The denoising loss is used to learn the new components, while the original weights of the pre-trained model are frozen. 
We will describe the technical details below. 

\subsection{Model Architecture Details}
Our approach is composed of several key components. 

\noindent\textbf{Data Pre-Processing.}
We use $X_t=\{x_t^i\}_1^N$ to represent the set of the original images, where $N$ is the number of input images. Since the object of the input concept in the images may not be large enough, we crop out the object from each image to obtain a set of conditional image $X_s=\{x_s^i\}_1^N$. To further enforce the model to focus on the exact object, we mask out the background of each cropped image. We have:

\begin{equation}
    x_s^i \coloneqq x_s^i \cdot m_s^i, 
\end{equation}
where $m_s^i$ is the mask of the object in the $i$-th  cropped image with $1$ indicating the pixel of object and $0$ indicating the background pixel. During training, we perform random augmentations to the masked image. We have $x_s^i \coloneqq \mathcal{A}(x_s^i) $, where $\mathcal{A}$ is the augmentation operator. The masked image set $X_s$ will be the final image condition to our model. 

\noindent\textbf{Prompt Creation.}
We create our prompts using a format of ``... $\hat{V}$ [class noun] ...'', where $\hat{V}$ is a unique identifier to represent the input subject, and [class noun] is a coarse category of the subject. Given the original prompt $p$, we obtain the modified prompt $p_s$ by putting the identifier right before the class noun of the object. For example, if the original prompt is ``A photo of a person playing guitar'', then the modified prompt with the identifier is ``A photo of a $\hat{V}$ person playing guitar'', where ``person'' is the class noun.  During training, for each object category, we first detect the corresponding nouns in the original prompt, then insert the identifier. For example, for ``person'' object category, we insert the identifier to the nouns that are coarse descriptions of ``person'', including ``man'', ``woman'', ``baby'', ``girl'', ``boy'', ``lady'', \etc. For ``cat'' category, we insert the identifier to the nouns ``cat'', ``kitten'' and so on.

\noindent\textbf{Concept Embedding Learning.} 
Having created the modified prompt, the next step is to encode the  concept into the model. To this end, we convert the input images into a textual concept embedding. Since the identifier $\hat{V}$ has indicated the location of the textual embedding, we adopt an image encoder $E_c$ to map the images to a compact concept feature vector $\textbf{f}_c$ in the textual space. Specifically, $\textbf{f}_c$ is the average feature vector of the global features of all input images. We have: 

\begin{equation}
    \textbf{f}_c = \sum_{i=1}^{N}E_c(x_s^i) / N.
\end{equation}

To obtain the final textual embeddings of the input prompt, we first obtain the CLIP~\cite{radford2021learning} Text embeddings $\textbf{c}_s$ of the modified prompt $\textbf{c}_s = CLIP(p_s)$, then replace the embedding of identifier $\hat{V}$ with the concept feature $\textbf{f}_c$ to obtain the concept injected textual embedding $c$. This final embedding will be the condition in the cross-attention layers of the text-to-image diffusion model. 

\noindent\textbf{Rich Representation Learning with Adapters.}
The learned concept embedding is a compact feature containing the global semantics of the input images since we need to encode the feature into the textual space. Consequently, the concept embedding may miss the fine details of the input subject, such as the shape of object parts, textural details, structure details, and so on. To better preserve the identity, one way is to inject rich features containing identity details into the pre-trained diffusion model. 

Our goal is to extend the pre-trained model to take additional visual features but retain the visual synthesis and language understanding abilities of the pre-trained large models. To this end,  we propose to freeze the original weights of the pre-trained model and extend the model with new trainable adapter layers. Specifically, the U-Net of the pre-trained diffusion model is composed of several Transformer blocks. The original Transformer block contains a self-attention layer followed by a cross-attention layer that takes both the visual feature tokens and the textual embeddings as inputs for cross-attention learning. In each Transformer block, we add a new learnable adapter layer between these two frozen layers as illustrated in Fig.~\ref{fig:framework}. The new layer is formulated as:

\begin{equation}
\label{eq_adapter}
    \textbf{y} \coloneqq \textbf{y} + \beta \cdot \tanh(\gamma) \cdot S([\textbf{y}, \textbf{f}_p]),
\end{equation}
where $S$ is the self-attention operator, $\textbf{y}$ is the visual feature tokens, $\gamma$ is a learnable scalar initialized as zero, $\beta$ is a constant to balance the importance of the adapter layer. $\textbf{f}_p$ represents the rich patch token sequences extracted using an image encoder $E_p$ from input images $X_s$. Specifically, for each image, we extract $257$ visual tokens with the patch encoder $E_p$.  Then we obtain the rich patch feature as a token sequence by concatenating the tokens from all the input images. We have:

\begin{equation}
    \textbf{f}_p = Concat(\{E_p(x_s^i)\}_1^N).
\end{equation}

Compared to the compact concept embedding, the patch feature contains subject-related content details of the input images; thus they can significantly benefit identity preservation.

Note that GLIGEN~\cite{li2023gligen} also adopts a similar adapter structure to take conditions. The primary difference from our work is that when GLIGEN takes an image as input, a global feature is extracted as the condition so that the image provides style or high-level content guidance, while we extract rich patch features (represented by a token sequence) from images for identity preservation. Moreover, we use a different sampling schedule during inference as we will depict below. 

\subsection{Model Training}
During training, we use heavy augmentation $\mathcal{A}$ to obtain variations of masked images $X_s$. The original image set $X_t$ (without cropping out the object region or masking out the background) is regarded as the ground-truth. Since we do not have paired images of the same concept as training data, we simply use $1$ image to train our model for each concept, i.e., $N=1$ in the image set. Our loss function to optimize the model is formulated as:

\begin{equation}
    \mathcal{L} = \mathbb{E}_{z,t,c,X_s,\eta \in \mathcal{N}(0,1)}\left[\|\eta - \eta_{\theta}(z_t, t, c, X_s)\|_2^2\right],
\end{equation}
where $z_t$ is the latent noisy image at time step $t$ obtained from the ground-truth image $x_t^1$, $\eta$ is the latent noise to predict, $c$ is the textual embedding, $X_s$ is the set of conditioning images, and $\eta_{\theta}$ is the noise prediction model with parameters $\theta$.

Only the adapter layers and the fully-connected layer of our image encoders are updated. The original weights of the pre-trained text-to-image model are frozen in the full training procedure.

\begin{figure}[t]
    \centering
    \includegraphics[width=0.46\textwidth]{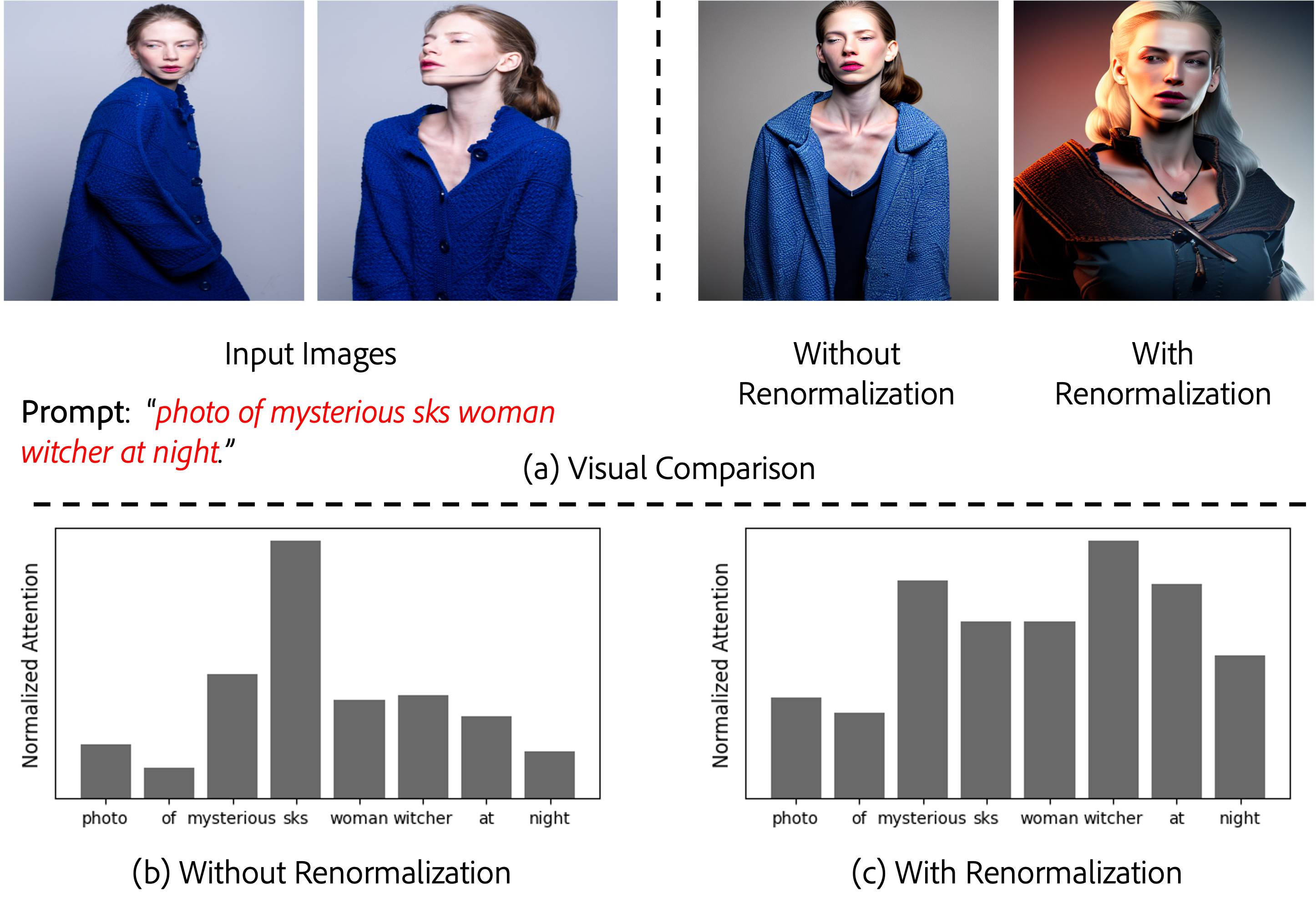}
    \caption{(a): Visual comparison between our model with and without concept renormalization. (b) The average attention of each word without concept renormalization. (c) The average attention of each word with concept renormalization.}
    
    \label{fig:attn}
\end{figure}

\begin{figure*}[t]
\vspace{-0.2in}
    \centering
    \includegraphics[width=\textwidth]{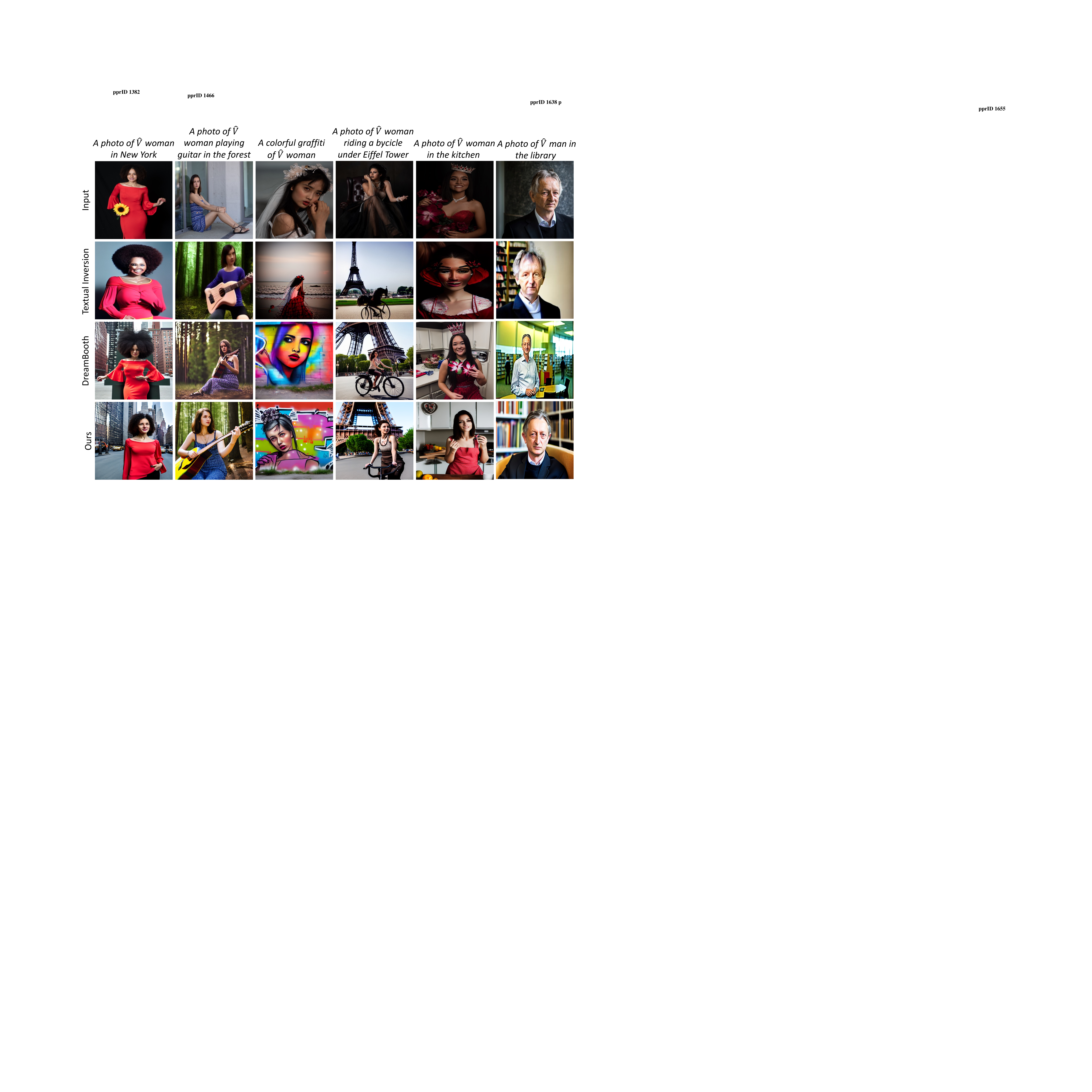}
    \caption{The visual comparison with other methods. All methods take five images as input but we only show one image here for simplicity.}
    \label{fig:main_comp}
    \vspace{-0.1in}
\end{figure*}

\subsection{Model Inference}

\noindent\textbf{Arbitrary Number of Input Images.}
During model's inference, we still mask out the background of the cropped images, but do not perform any augmentations to the masked images, i.e., $\mathcal{A}=None$. Although the rich patch feature $\textbf{f}_p$ is composed of tokens from only one image during training,  our adapter can actually take arbitrary number of conditioning images as inputs during inference. This flexibility is owning to the concatenation operation in the adapter layer and the nature of self-attention. 

\noindent\textbf{Balanced Sampling.}
During training, $\beta$ in Eq. \ref{eq_adapter} is set to $1$. During inference, however, we observe that setting $\beta$ to $1$ results in a strong reconstruction of the input images with good identity preservation, while the language-image alignment is weakened. Since the original pre-trained model has a deep understanding of the language, we reduce the value of $\beta$ during inference so that the adapter layer takes both the visual information from the original pre-trained model and the conditioning images. We observe that $\beta$ actually plays the primary role for achieving a good balance between language understanding and identity preservation. More analysis on the influence of this parameter is present on the experiment section. 

\noindent\textbf{Concept Token Renormalization.}
We perform further analysis on the language understanding of our model. We observe that even if we have adjusted the value of $\beta$ for balanced sampling, the concept token for the identifier $\hat{V}$ can sometimes dominate the cross-attention between visual tokens and textual tokens in the Cross-Attention layers, leading to language forgetting. Fig. \ref{fig:attn} (b) shows the cross-attention averaged over all batches, layers, and time-steps for each word when inputting the prompt "photo of mysterious sks woman witcher at night", where ``sks'' denotes the identifier $\hat{V}$. The adapter weight $\beta$ is set to 0.3 in this case. The attention of the identifier is significantly higher than the other words, while the key words such as ``night'' and ``witcher'' are assigned low attention weights, showing a sign of language forgetting. 

To address this issue, we renormalize the concept token with a factor of $\alpha \in (0, 1]$. We have: 

\begin{equation}
   \textbf{f}_c = \alpha \cdot  \textbf{f}_c.
\end{equation}

Since there are only linear mappings from the prompt embeddings before calculating the cross-attention, such renormzation stretagy is essentially equivalent to rescaling the cross-attention between the concept token and the visual tokens in the Cross-Attention layers. Fig. \ref{fig:attn} (c) show the average attention for each word after our renormalization. It can be observed that the attention of the concept token does not dominate the cross-attention anymore. The visual comparison shown in Fig. \ref{fig:attn} (a) provides more evidence for this observation. Without concept renormalization, the model failed to generate the ``witcher'' style or the ``night'' background. With renormalization, the attentions of the nouns are more balanced, and the model successfully generates the ``witcher'' style and the ``night'' background. All the analysis and evidence demonstrate that our concept renormalization can help to achieve a better balance between text-image alignment and identity preservation. 

\begin{figure*}[t]
    \centering
    \includegraphics[width=\textwidth]{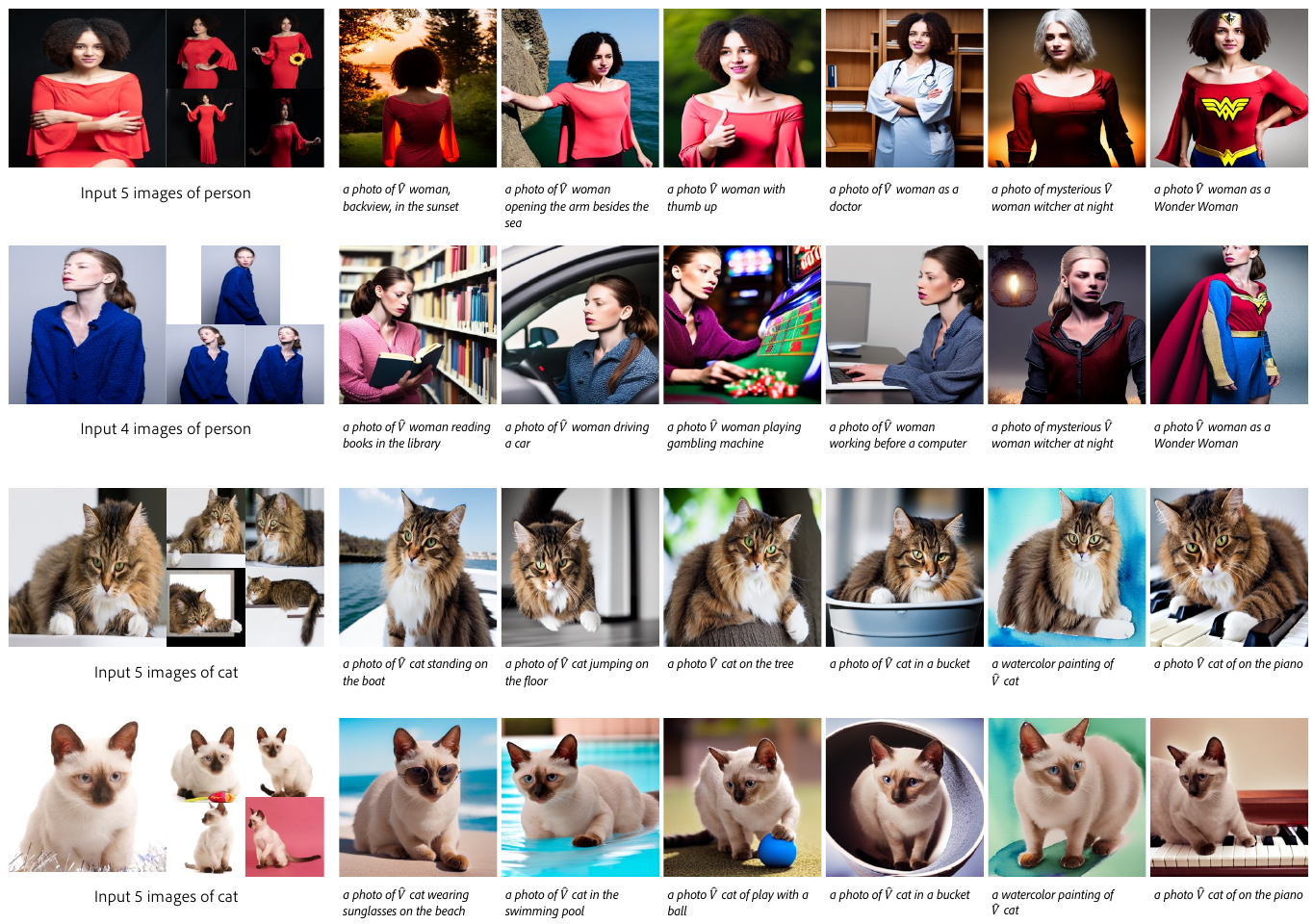}
    \caption{Personalized images generated by our model on the ``\textit{person}'' and ``\textit{cat}'' categories.}
    \label{fig:visual}
    \label{fig:more_visual}
\end{figure*}

\section{Experiment}
\subsection{Dataset and Metric}
\noindent\textbf{Datasets}. We conduct experiments on two subject categories \textit{person} and \textit{cat}, and collect text-image pairs where the images contain these object categories and the prompts contain the related coarse category nouns.  The coarse descriptions of the \textit{person} category includes ``person'',  ``man'', ``woman'', ``baby'', ``girl'', ``boy'', ``lady'', etc. The coarse descriptions of the \textit{cat} category includes the nouns ``cat'', ``kitten'' and so on.  Then we collect the entity segmentation masks for all the collected images using the entity segmentation model~\cite{9976289} trained on high-quality entity segmentation datasets~\cite{qi2022fine}. Then we filter out images where the region ratio of the object belonging to our target subject is less than 0.1 or larger than 0.7. We also filter out images with multiple objects to simplify the training. 
In total, we have 1.43 million text-image pairs for person category and 0.37 million for cat category.
We leverage PPR10K~\cite{liang2021ppr10k} dataset as our testing dataset for person category. PPR10K includes high-quality human portrait photos of 1681 identities, each of which contains multiple images of the same person. We select 50 identity in the test split of PPR10k~\cite{liang2021ppr10k}, where each selected identity is guaranteed to have more than 5 images and we only keep the first 5 images in naming order as our test input.

\noindent\textbf{Metrics}.
We quantify the identity preservation ability and vision-language alignment of our model using the following metrics.\\
\noindent \textit{Reconstruction} is to evaluate whether the identity can be fully preserved by the default prompt ``A photo of $\hat{V}$ [class noun]'', where [class noun] can be person or cat. It is measured by the similarity of CLIP visual features between the input image and the generated image.\\
\noindent\textit{Face distance} is to especially evaluate whether the identity can be fully preserved for the ``person'' category with various prompts. As the face is the core indicator of the identity, we use a strong face detector \cite{serengil2021lightface} to detect the faces in both the generated and the input images of a subject. Then we extract an embedding from each detected face with an Inception-ResnetV1~\cite{szegedy2017inception} pre-trained on VGGFace2~\cite{cao2018vggface2}. Then we calculate the average embedding distance of each pair of faces as the perceptual face distance. Please refer to Appx.~\ref{appx:face_metric} for more details.\\
\noindent\textit{Alignment} is to measure the vision-language alignment between the input prompt and the output image, indicating whether the generated image follows the prompt.
We use the CLIP similarity between image and text embeddings to measure it. 
We construct various prompts ranging from background modifications (``A photo of $\hat{V}$ [class noun] on the moon''), to style changes (``An oil painting of $\hat{V}$ [class noun]''), and a compositional prompt (``$\hat{V}$ [class noun] shaking hand with Biden''). The specific prompt list is in Appx.~\ref{appx:prompt}.

\subsection{Implementation Details}
We utilize the Stable Diffusion~\cite{rombach2022high} V1-4 as our pre-trained text-to-image model, which is the current leading model available to the public. For all experiments of our model, we use ``sks'' as the unique identifier $\hat{V}$ as suggested in DreamBooth~\cite{dreamboothsd}. For both the concept encoder $E_c$ and the patch encoder $E_p$, we use the pre-trained CLIP image encoder as the backbone followed by a randomly initialized fully-connected layer. During training, we freeze the backbone of the image encoders and only update the FC layers and the adapter layers. The weights of CLIP text encoder and the original weights in the U-Net of the pre-trained text-to-image model are also frozen. 
Our model is trained for 320k iterations for person and 200k iterations for cat,  with the learning rate 1e-6 for adapter layers and 1e-4 for the FC layers in the visual encoders, under batch size 16 deployed over 4 A100 GPUs.

\subsection{Comparison to SOTA Methods}
In this section, we compare our approach with Textual Inversion~\cite{gal2022image} and DreamBooth~\cite{ruiz2022dreambooth}, both requiring heavy test-time finetuning. We adopt the official code base for Textual Inversion and third-party replication\cite{dreamboothsd} for DreamBooth where the base text-to-image model is replaced from Imagen~\cite{saharia2022photorealistic} to Stable Diffusion~\cite{rombach2022high}.


\noindent\textbf{Qualitative Results.}
The personalization results on both person and cat categories are shown in Fig.~\ref{fig:teasing}, and the comparison results with other methods are shown in Fig.~\ref{fig:main_comp}, where our method exhibits better perceptual quality, vision-language alignment and identity preservation ability than the compared ones.
We observe that our method can also support large pose and structure variations, such as ``riding bycicle'' and ``open arms''.

For Textual Inversion, its text-to-image model is LDM~\cite{rombach2022high}, which has weaker capacity than Stable Diffusion. Therefore, it tends to yield blurry images and cannot correctly follow the prompts.
DreamBooth can generate high-quality images. However, for the input image with smaller portion of the person object (e.g. column 2 and 4 of Fig.~\ref{fig:main_comp}), the generated images also tends have small portion of object and thus the identity preservation ability is limited.
Moreover, even if the input image contains a large portion of the person object (e.g. column 5 of Fig.~\ref{fig:main_comp}), DreamBooth can only preserve the person's outfit but still distort the face identity.
In contrast, our method can generate images with clearer faces and details given a wide range of person size portion in the image. We suspect the reason is that our adapter layers have seen millions of different person identities; therefore it garners stronger prior for identity keeping than the compared test-time finetuning-based methods.

Fig.~\ref{fig:more_visual} shows more visual results generated by our model. On the ``\textit{person}'' and ``\textit{cat}'' categories, our model can both generate diverse, identity-preserved and language-aligned images. 

\begin{figure}[t]
    \centering
    \includegraphics[width=0.48\textwidth]{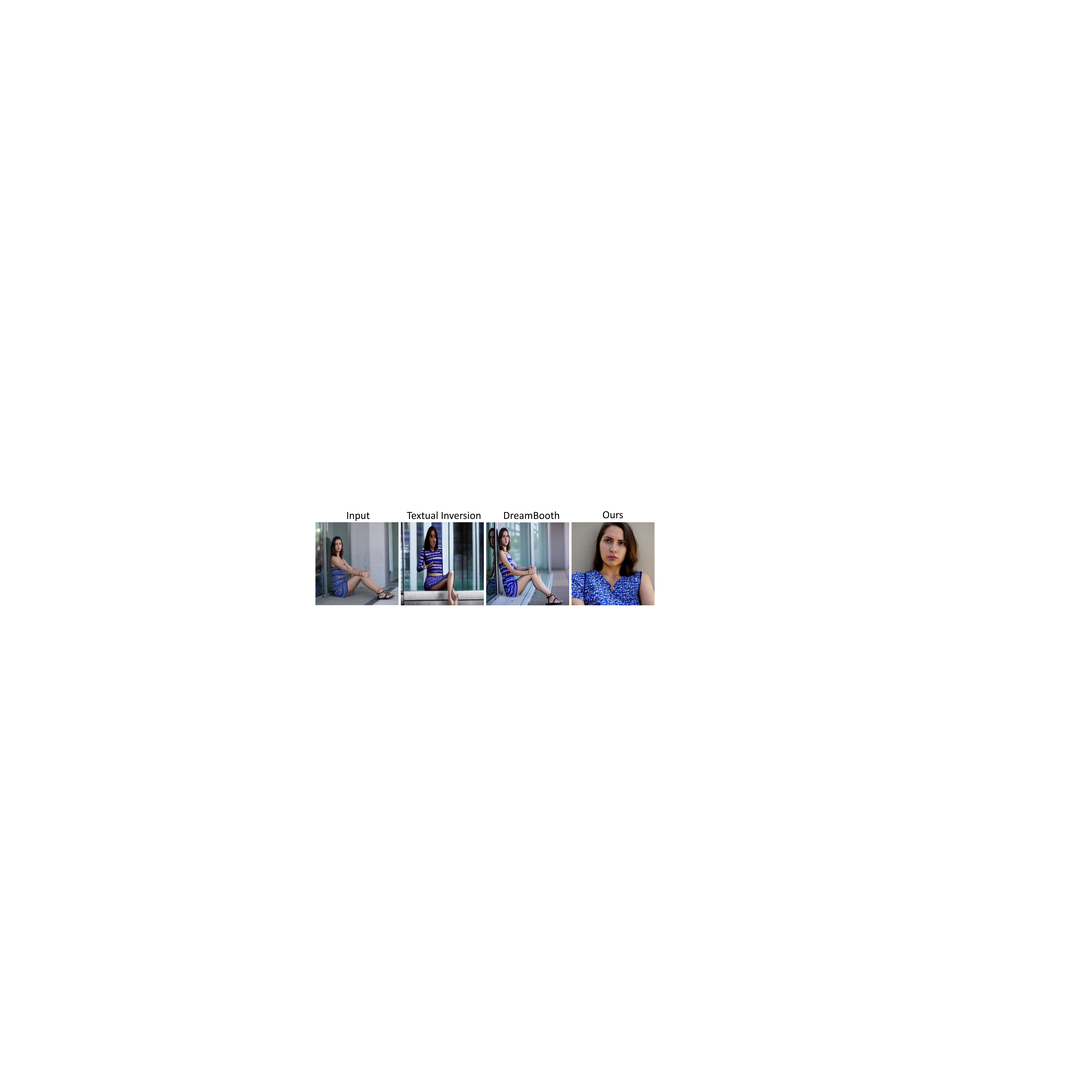}
    \caption{The visual comparison of image reconstruction given prompt ``A photo of $\hat{V}$ woman''.}
    \label{fig:reconstruct}
\end{figure}

\noindent\textbf{Quantitative Results.}
The quantitative comparison with other methods is shown in Tab.~\ref{tab:main_comp}.
Our method achieves better vision-language alignment and face distance than the comparison methods, but the reconstruction is weaker, which can be explained via the visualization of the reconstruction in Fig.~\ref{fig:reconstruct}.
Both Textual Inversion and DreamBooth tend to replicate both the foreground person and the background glass wall, while our method reconstructs only the portrait without background content.
The reason is that during training of our model, we add the mask to suppress the background content; therefore our model learns to primarily keep the identity of the foreground object, but not the background.
This background discrepancy leads to a lower reconstruction score of our method, but does not necessarily mean our method is inferior in identity preservation.
Therefore, although DreamBooth and Textual Inversion focus more on reconstructing the full image during finetuning, our model can generate faces that are significantly more similar than the other methods. 

To test of our model in full setting (i.e., the inputs are background-masked images during test), we also calculate the alignment score and the perceptual face distance for our model with masked images as input. The results in Tab. \ref{tab:main_comp} (last row) demonstrate that both our model variants achieve better vision-language alignment and identity preservation. 

Moreover, since our method dose not need test-time finetuning, our testing time in Tab.~\ref{tab:main_comp} (last column) is significantly lower than the other methods (100x faster).

\begin{figure*}[t]
    \centering
    \includegraphics[width=\textwidth]{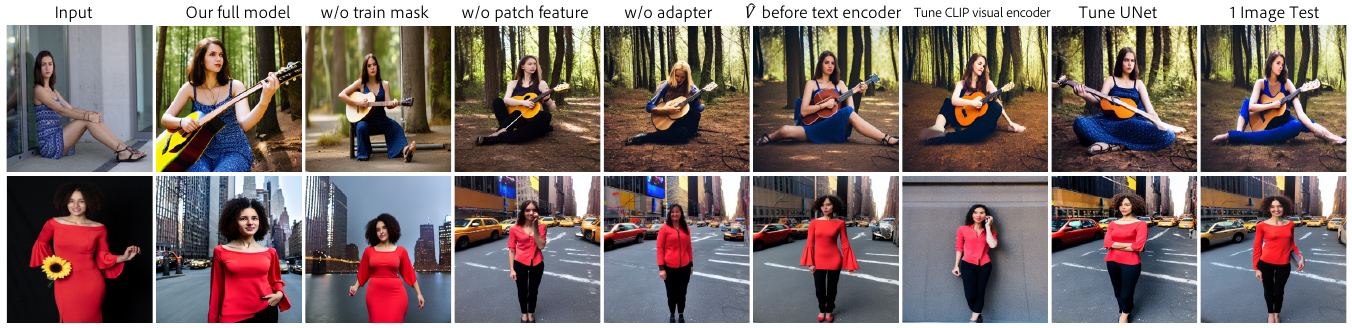}
    \caption{The visual comparison of different ablation settings. The random noise is the same for all variants of our model. The prompts and input images are taken from Fig.~\ref{fig:main_comp}.}
    \label{fig:main_abla}
\end{figure*}


\begin{table}[t]\centering
\ra{1.2}
\scalebox{0.95}{
\begin{tabular}{@{}lrrrr@{}}
\toprule
Methods & Align $\uparrow$ & Face dist $\downarrow$ & Recon$\uparrow$ & Time (s) $\downarrow$\\
\midrule
TI~\cite{gal2022image}  & 0.2556  & 1.5462 & 0.7832  &  $\sim$ 1500 \\
DB~\cite{ruiz2022dreambooth} & 0.3088  & 1.2281 & \textbf{0.8335} &  $\sim$ 600 \\
Ours & \textbf{0.3140}  & 1.1901 & 0.7329 & \textbf{6}\\
Ours + M & 0.3135  & \textbf{1.1899} & - & \textbf{6}\\
\bottomrule
\end{tabular}}
\vspace{2mm}
\caption{Quantitative comparison in ``\textit{person}'' category of TI (Textual Inversion), DB (DreamBooth) and our method. The metric ``Align'' is for alignment, ``Face dist'' for face distance and ``Recon'' for reconstruction. ``M'' denotes to our model tested with masked images as input.}
\label{tab:main_comp}
\end{table}

\begin{table}[t]\centering
\ra{1.2}
\scalebox{0.95}{
\begin{tabular}{@{}lrrrr@{}}
\toprule
Methods & Quality $\uparrow$ & Alignment$\uparrow$ & Identity $\uparrow$\\
\midrule
Textual Inversion &  2.69 & 2.72 & 2.70 \\
DreamBooth & 3.55 & 3.44 & 3.48 \\
Ours & \textbf{3.89} & \textbf{3.72} & \textbf{3.56}  \\
\bottomrule
\end{tabular}}
\vspace{2mm}
\caption{User study on the ``\textit{person}'' category. ``Quality'' measures the image quality (\eg artifact-free), ``Alignment'' measures the vision-language alignment, and ``Identity'' measures the identity preservation performance.}
\label{tab:user_study}
\end{table}

\noindent\textbf{User Study.}
We conduct a user study to compare our method with DreamBooth and Textual Inversion perceptually.
For each evaluation, each user will see one input image, one prompt, and three images generated by each method. 
The user will rank each generated image from $1$ (worst) to $5$ (best) concerning its visual quality, vision-language alignment, and identity preservation.
We select $5$ identities of the ``person'' category where each identity is personalized by $10$ prompts.
$4$ generated images per prompt are evaluated, resulting in $200$ unique evaluation samples.
The user study is deployed via Amazon Mechanical Turk, where each sample will be evaluated by two different users, leading to 400 evaluated samples. 
After filtering out invalid user inputs, we obtained 344 valid evaluated samples.
The results shown in Tab.~\ref{tab:user_study} indicate that our method outperforms the two comparison methods concerning all the three important aspects.

\subsection{Ablation Study}
We conduct a thorough ablation study across various components and settings as follows.

\noindent\textit{W./o. train mask}.
To demonstrate the importance of the object mask, we remove the object mask used during training.

\noindent\textit{W./o. patch feature}.
To verify the necessity of rich patch feature as the condition to the model, we do not use the patch feature and only use the global [CLS] token feature from the CLIP viusal encoder as the input to the adapter layers.

\noindent\textit{W./o. adapter}.
We also remove the adapter branch and study if purely $\hat{V}$ can bear enough identity information.

\noindent\textit{$\hat{V}$ before text encoder}. 
Since our model inserts the $\hat{V}$ to the textual space after the CLIP text encoder, we also study the early integration of textual and visual information, i.e., we insert $\hat{V}$ into the text token space before CLIP text encoder.

\noindent\textit{Tune CLIP visual encoder}.
Our standard setting freezes both the backbone of CLIP visual encoder and only finetunes the FC heads.
However, since the adapter is designed to extract more fine-grained content details, we also investigate whether finetuning the backbone of CLIP visual encoder can benefit the learning of object details.

\noindent\textit{Tune U-Net}.
Inspired by DreamBooth where the U-Net of the pre-trained model is also tuned, we also try to tune the U-Net parameters.

\noindent\textit{Single image as input}.
Since our model is flexible for the number of input images, we evaluate our model using a single image as the input image condition, i.e., $N=1$.

\begin{table}[t]\centering
\ra{1.2}
\scalebox{0.95}{
\begin{tabular}{@{}lrrrr@{}}
\toprule
Methods & Align $\uparrow$ & Reconstruct$\uparrow$\\
\midrule
InstantBooth & 0.3140 & 0.7329\\
\midrule
w/o train mask & 0.3127 & 0.7485 \\
w/o patch feature & 0.3269 & 0.6494\\
w/o adapter & 0.3242 & 0.5468 \\
$\hat V$ before CLIP & 0.3127 & 0.7495\\
Tune CLIP Vis Enc & 0.3266 & 0.6425\\
Tune U-Net & 0.3142 & 0.7265 \\
1 Image Test & 0.3140 & 0.7261 \\
\bottomrule
\end{tabular}}
\vspace{2mm}
\caption{Ablation study for various settings of our model.}
\label{tab:main_abla}
\end{table}

\begin{table}[t]\centering
\ra{1.2}
\scalebox{0.9}{
\begin{tabular}{@{}ccccc@{}}
\toprule
$\beta$ & $\alpha$ & Align $\uparrow$ & Reconstruct$\uparrow$\\
\midrule
0.3 & 0.1 & 0.3242 & 0.6631 \\
0.3 & 0.3 & 0.3232 & 0.7002 \\
0.3 & 0.4 & 0.3140 & 0.7329 \\
0.3 & 0.5 & 0.3032 & 0.7544 \\
0.3 & 1.0 & 0.2087 & 0.7905 \\
0.5 & 0.1 & 0.3127 & 0.7051 \\
0.5 & 0.3 & 0.3076 & 0.7480 \\
0.5 & 0.5 & 0.2874 & 0.7778\\
1.0 & 1.0 & 0.2017 & 0.7944\\
\bottomrule
\end{tabular}}
\caption{Hyper-parameter adjustment for adapter weight $\beta$ and concept renormalization factor $\alpha$.}
\label{tab:main_hyper}
\end{table}

The quantitative results are present in Tab.~\ref{tab:main_abla} and the visual results are shown in Fig.~\ref{fig:main_abla}.
We observe that all the ablation settings result in weaker visual results than our full setting.
Without the object mask during training, the model fails to capture the accurate foreground information and the generation ability and language-understanding ability of the our model will be jeopardized by the background noise.
Therefore, the model tends to keep more background information, and thus has a higher reconstruction score but a lower alignment score.

Without the rich patch feature, the text-image alignment becomes slightly better but reconstruction degrades heavily, which means that the patch feature is crucial to capture the foreground detail and keep the identity. Similar analysis applied to the whole adapter branch.

For $\hat V$ before CLIP, we observe that there is a slight trade-off between alignment and reconstruction, but the visual results become worse. We conjecture that since the CLIP text encoder is frozen, the identity information from $\hat V$ is diffused by the CLIP text encoder and thus the visual details are missing. 

The results of tuning the CLIP visual encoder or U-Net indicate that tuning these modules will lead to worse identity preservation. Therefore, it is important to keep the weights of the pre-trained model fixed so that the original model capacity is preserved. 

The single image finetuning version of our model achieves the same alignment score but worse reconstruction, indicating that more input images can provide more details of the foreground to better preserve the identity. Nonetheless, from Fig.~\ref{fig:main_abla}, it is impressive that our model can achieve promising identity preservation even with one input image.

\noindent\textbf{Adjust the adapter weight $\beta$ and concept renormalization factor $\alpha$}. 
Tab.~\ref{tab:main_hyper} shows different compositions of $\beta$ and $\alpha$.
The results indicate that larger $\beta$ or $\alpha$ can both contribute to better identity preservation but weaker language comprehension ability. 
We finally choose the model with $\beta=0.3$, $\alpha=0.4$ as a trade-off.

\section{Conclusion}
We presented an approach that extends existing pre-trained text-to-image diffusion models for personalized image generation without test-time finetuning. The core idea is to convert input images to a textual token for general concept learning, and introduce adapter layers to adopt rich image representation for generating fine-grained identity details. Extensive results demonstrated that our model can generate language-aligned, identity-preserved images on unseen concepts with only a single forward pass. This remarkable efficiency improvement will enable a variety of practical personalization applications.

\noindent\textbf{Limitation and Future Work}
While our model exhibits strong performance and fast speed, it still has several limitations.
First, we have to separately train the model for each category. This limitation can be addressed by training the model with more data of different categories together.
In addition, due to the current design of adapter, it can only accept a single concept to provide the identity details.
We plan to address these limitations and extend our approach to personalized image editing and video generation as our future work.

\section*{Acknowledgement} We would like to thank Qing Liu for dataset preparation and He Zhang for object mask computation.

{\small
\bibliographystyle{ieee}
\bibliography{ms}
}
\clearpage

\newcommand{\appendixhead}%
{\centering\textbf{\huge Appendix}
\vspace{0.5in}}
\twocolumn[\appendixhead]

\appendix

\section{Evaluation on Identity Preservation with Perceptual Face Distance} \label{appx:face_metric}
Since during the testing phase, both DreamBooth and Textual Inversion directly resize the input images (which also serve as the training samples for these models) to the target size, for a fair comparison, we test our model without cropping or masking the object across all our experiments in this paper. We use the reconstruction score calculated between each pair of full generated and input images to evaluate identity preservation. However, a large portion of the background region exists in both the generated and the input images, significantly affecting the similarity comparison between the generated foreground object and the input object. Therefore, using the full image to calculate the reconstruction score is indeed not an accurate metric for evaluating identity preservation. 

To address this issue, for the ``\textit{person}'' category, we use a strong face detector \cite{serengil2021lightface} to detect the faces in both the generated and the input images of a subject. Then we extract an embedding from each detected face with an Inception-ResnetV1~\cite{szegedy2017inception} pre-trained on VGGFace2~\cite{cao2018vggface2}. Then we calculate the average embedding distance of each pair of faces as the perceptual face 
 distance. Specifically, we use the L-2 norm of the two embeddings as the distance between two faces.  Lower distance indicates the faces are more similar. The face detector successfully detects the faces in all the input images and most of the generated images. We checked some of the generated images where faces cannot be detected. The face detector only fails on generated faces that are extremely distorted, demonstrating that the pre-trained face detector is trustworthy.  


\section{Explanation on The Necessity of Using The $\hat{V}$ Identifier}
\begin{figure}[t]
    \centering
    \includegraphics[width=0.47\textwidth]{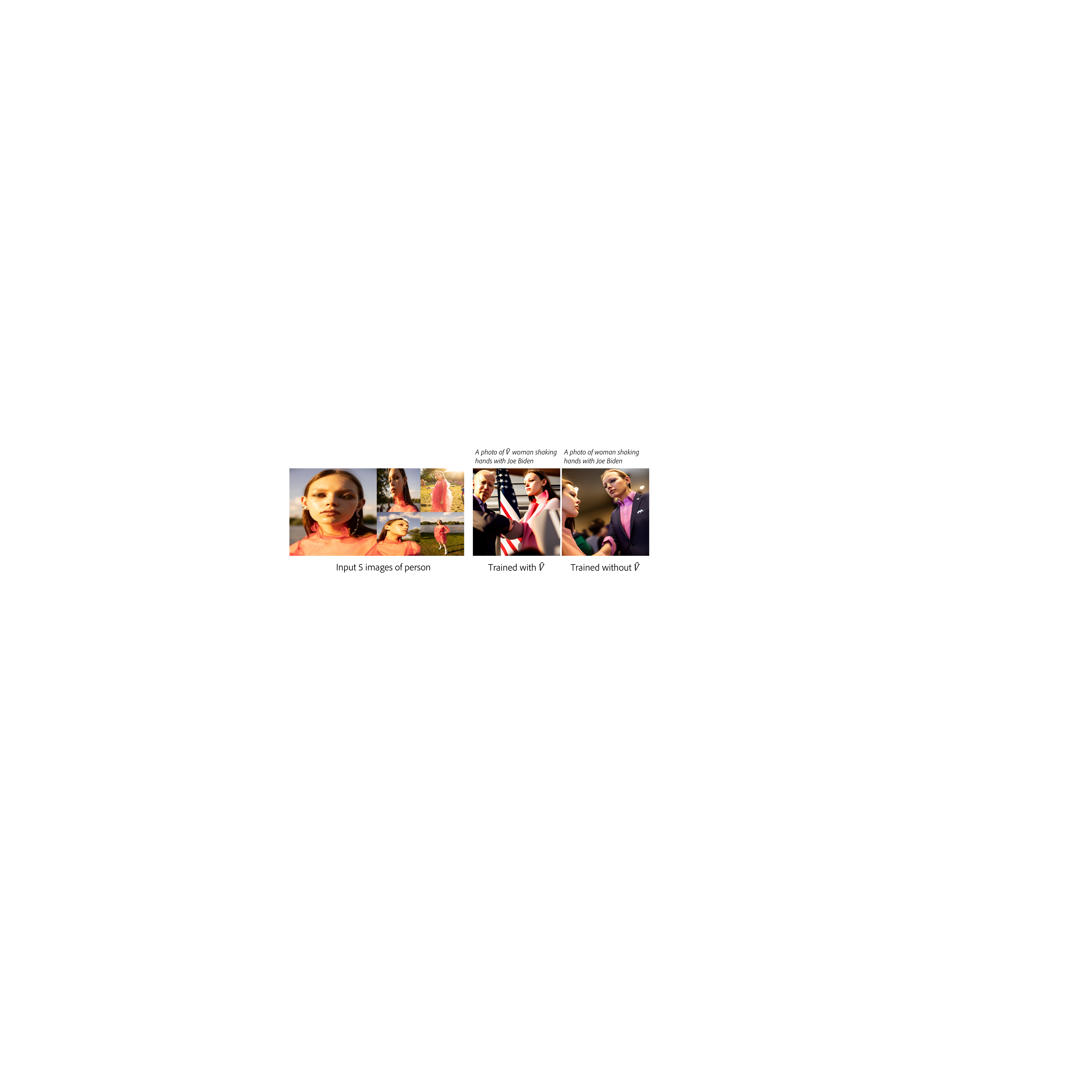}
    \caption{Visual comparison between our full model trained with the $\hat{V}$ identifier and our model variant trained without using the $\hat{V}$ identifier.}
    \label{fig:sks_cmp}
\end{figure}

This section investigates the importance of using the identifier $\hat{V}$ in the input prompt. We visually compare our model trained with $\hat{V}$ identifier and the variant of our model trained without $\hat{V}$. Results in Fig. \ref{fig:sks_cmp} show that the model cannot generate Joe Biden's face correctly without training with $\hat{V}$. Interestingly, Joe Biden's face is similar to the input woman's. This observation indicates that the model is confused about which object in the generated image should be aligned with the input images. On the contrary, our model trained with $\hat{V}$ identifier can successfully generate the face of Joe Biden. The comparison demonstrates the severe drawback of not using the identifier in the prompt. We conclude that using the $\hat{V}$ identifier is necessary to guide our model to align the input images to the correct subject in the prompt and the generated images.  



\section{Details of Test Prompts} \label{appx:prompt}
The specific prompts that we use to obtain the quantitative results for person are as follows.

\noindent\textit{``a photo of $\hat{V}$ [class noun] in the swimming pool'' \\
``a photo of $\hat{V}$ [class noun] in New York''\\
``a photo of $\hat{V}$ [class noun] on the moon''\\
``a photo of $\hat{V}$ [class noun] in the kitchen''\\
``a photo of $\hat{V}$ [class noun] in mountain with aurora''\\
``a photo of $\hat{V}$ [class noun] in the library''\\
``a painting of $\hat{V}$ [class noun] in Van Gogh style''\\
``a manga drawing of $\hat{V}$ [class noun]''\\
``a colorful graffiti of $\hat{V}$ [class noun]''\\
``a $\hat{V}$ [class noun] funko pop''\\
``a pencil drawing of $\hat{V}$ [class noun]''\\
``a Ukiyo-e painting of $\hat{V}$ [class noun]''\\
``a photo of $\hat{V}$ [class noun] holding a corgi on a bench''\\
``a photo of $\hat{V}$ [class noun] playing guitar in the forest''\\
``a photo of $\hat{V}$ [class noun] shaking hands with Joe Biden''\\
``a photo of mysterious $\hat{V}$ [class noun] witcher at night''\\
``a photo of $\hat{V}$ [class noun] riding a bycicle under Eiffel Tower''\\
}

\end{document}